\documentclass{article}

\usepackage{agi-10}

\nocopyright

\usepackage{graphicx}
\usepackage{algorithm}
\usepackage{algorithmic}
\usepackage{amsmath}
\usepackage{amstext}
\usepackage{amsfonts}
\usepackage{amsbsy}
\usepackage{amsthm}
\usepackage{prettyref}

\newtheorem{lemma}{Lemma}

\newcommand{\fancy}[1]{\mathcal{#1}}
\newcommand{\F}{\fancy{F}}

\newcommand{\algo}[1]{\textsc{#1}}

\newrefformat{alg}{Algorithm~\ref{#1}}
\newrefformat{eq}{Equation~\ref{#1}}
\newrefformat{lem}{Lemma~\ref{#1}}
\newrefformat{thm}{Theorem~\ref{#1}}
\newrefformat{chp}{Chapter~\ref{#1}}
\newrefformat{sec}{Section~\ref{#1}}
\newrefformat{apx}{Appendix~\ref{#1}}
\newrefformat{tab}{Table~\ref{#1}}
\newrefformat{fig}{Figure~\ref{#1}}

\title{Abstract Representations and Frequent Pattern Discovery}
\author{Eray \"Ozkural \\
Bilkent University Computer Engineering Department \\
erayo@cs.bilkent.edu.tr}

\begin{document}

\maketitle

\begin{abstract}
  We discuss the frequent pattern mining problem in a general setting. From an
  analysis of abstract representations, summarization and frequent pattern mining,
  we arrive at a generalization of the problem. Then, we show how the
  problem can be cast into the powerful language of algorithmic information
  theory. This allows us to formulate a simple algorithm to mine for all
  frequent patterns. \footnote{: This paper has been written in 2006, however, it was
  generously refused from two conferences; one of them being ICDM. And the
  later one is AGI-10. The paper is going to be revised, eventually,
  according to the AGI-10 reviewers' criticism, which 
  claimed, among other things, that algorithmic statistics supersedes
  this paper. The author would like to share this paper with the
  public until the criticism can be adequately addressed,
  which will unfortunately not be possible for some time due to my PhD studies.}
\end{abstract}

\section{Introduction }
\label{sec:intro}

The field of data mining is changing faster than we can define it. In recent
years, foundations of data mining has received considerable interest,
helping remove some of the ad-hoc considerations in the theory of data
mining and expanding the frontiers. The problem definitions of early data
mining research have now been analyzed meticulously, considering especially
the performance and scalability of methods, giving a performance-oriented
character to most data mining research. Qualitative work has usually focused
on slight variations of the original problems; staying within the framework
of basic problems such as association rule mining and sequence mining.
However, the ever expanding computational and storage capacity challenges us
to devise new ways to look at the data mining tasks, to discover more
interesting/useful patterns. The subject of this paper is a substantial
revision of the frequency mining problem, this time mining for any kind of a
pattern instead of frequent item sets. We arrive at our formulation from a
philosophical analysis of the problem, conceiving what the problem might
look like in the most general setting. After reviewing some of the recent
literature on generalizing data mining problems, we examine the relation of
abstraction to the summarization task and in particular frequent pattern
discovery. We then present a novel formulation of the frequent pattern
discovery problem using algorithmic information theory, derived from our
philosophical analysis. We show that our formulation exhibits similar formal
relations to the original frequent itemset mining problem, and is arguably a
good generalization of it. Then, we present the \algo{Micro-Synthetic}
algorithm which has the capability to detect any kind of a pattern given our
information theoretic definition of pattern occurrence. The algorithm is
similar to the \algo{Apriori} algorithm in its logic of managing the task in
a small number of database scans. After discussing the pros and cons of our
approach, we outline future research directions.

\section{Related work}
\label{sec:related}

We will skip the definitions and methods of traditional frequent pattern
discovery for considerations of space. For an introduction to the subject,
see \cite{agrawal94fast,zaki99parallel,hipp00}. There has been some
promising research in applying the generic methods of Kolmogorov complexity
to data mining. The authors report favorable results for classification and
deviation detection tasks in \cite{paramfree}. A mathematical theory of high
frequency patterns which uses granular computing was presented in
\cite{lintheory}. We will now take a closer look at algorithmic methods
which have attracted a great deal of interest.


\subsection{Algorithmic Information Theory}

Algorithmic information theory (AIT) gives an absolute characterization of
complexity for arbitrary bit strings~\cite{ait}. A computer is a computable
partial function $C(p,q)$ of self-delimiting program strings $p$ and data
$q$, where both input and output datum are bitstrings in $\{0,1\}^*$. Empty
string is denoted with $\Lambda$ and the shortest program which computes $s$ is
denoted with $s*$. $U$ is a universal computer that can simulate any other
computer $C$ with $U(p',q) = C(p,q)$ and $ |p'| \leq |p| + sim(C)$ where
$sim(C)$ is the length of simulation program for $C$. An admissible
universal computer is LISP with its eval function.

The algorithmic information content $H(s)$ of a bit string $s$ is the size
of minimal program $s*$ which computes it.  $H(s/t)$ is the algorithmic
information content of $s$ relative to $t$ (conditional algorithmic
entropy). Another definition from AIT is mutual algorithmic information
$H(s:t)$ which is relevant to our work. $H(s:t)$ is the extent to which
knowing $s$ helps one to calculate $t$.  The probability $P(s)$ of a
bitstring $s$ is the probability a program evaluates to $s$. Likewise, the
conditional probability $P(s/t)$ is the probability a program evaluates to
$s$ given the minimal program $t*$ for calculating $t$.
 In the definitions below, the
characterizations are given for particular computers, when we use a
universal computer for $C$ we drop the subscript (Since then the definitions
indeed become universal in that the difference between a universal computer
and another computer will always be a constant simulation cost, a result
which is known as the invariance theorem)

AIT gives an analogous formalism to information theory, and is deemed more
fundamental since Shannon information can be derived from algorithmic
(Kolmogorov) information. It is not possible to include all theorems here,
but some relevant consequences and results will be stated, mostly without
proof.

$H(s,t)$ is the joint algorithmic information of $s$ and $t$ where ``$,$''
denotes concatenation of bitstrings (at any rate it is straightforward to
convert between any two pair encodings). Algorithmic information is
asymptotically symmetric, e.g. $H(s,t)=H(t,s) + O(1)$ since in high level
languages it is not problematic to accomplish this sort of feat with a short
constant program. The conditional entropy of a string with itself is
constant, similarly.



Theorem I8 of \cite{ait} states that conditional entropy measures how easier
it is to compute two strings together than separately.
\begin{equation}
\label{thm:condentropy}
  H_C(t/s) = H(s,t) - H(s) + c
\end{equation}
 
Theorem I9 of \cite{ait} exposes the relationships between joint, mutual and
conditional information, as well as probability and joint probability. In
particular, algorithmic information is subadditive and conditional and
mutual information can be calculated from probabilities.
\begin{subequations}
\begin{align}
\label{thm:mutentropy}
  &H(s,t) = H(s) + H(t/s) + O(1)\\
  &H(s:t) = H(s) + H(t) - H(s,t) + O(1) \\
  &H(s:t) = H(t:s) + O(1) 
\end{align}
\end{subequations}

There are several other interesting theorems in AIT, however they fall
beyond the scope of the present work.

\subsection{Algorithmic distance metrics for classification}
\label{sec:ncd}

$H(s)$ is uncomputable. However, it can be approximated with a reasonable
compression program from the above. The standard UNIX compression programs
gzip and bzip2 have been used exactly for this purpose by
Cilibrasi~et~al~\cite{rudi-clustering} for clustering music files. In the
predecessors to this paper,
Vitanyi~et~al~\cite{information-distance,similarity-metric,rudi-clustering-compression} have introduced a distance function based on
algorithmic information theory which can be used for domain unspecific
classification and clustering algorithms.

Let $\rho(.,.)$ be a distance function for a metric space $(A,\rho)$, then the
following must hold:
\begin{enumerate}
\item $\rho(a,b) = 0$ if and only if $a=b$
\item $\rho(a,b) = \rho(b,a)$ (symmetry)
\item $\rho(a,c) \leq \rho(a,b) + \rho(b,c)$ (triangle equality)
\end{enumerate}

The authors propose to use normalized information distance (NID)  which is
based on algorithmic (Kolmogorov) complexity of bit strings.
\begin{equation}
  \label{eq:nid}
  \begin{split}
    &d : \{0,1\}^* \times \{0,1\}^* \to [0,1] \\
    &d(a,b) \mapsto \frac{max\{  H(a/b), H(b/a) \} }{max \{ H(a), H(b) \} }
  \end{split}
\end{equation}
If we assume $H(y)\geq H(x)$ without loss of generality, then we can
substitute $H(y/x) = H(y) - H(x:y) $
from basic definitions and we will have
\begin{equation*}
  d(x,y) = 1 - \frac{H(x:y)}{H(y)}
\end{equation*}
which satisfies requirements for a metric function.

A significant property of NID is the normalization of distance according
to the dominating complexity among two objects. Therefore, a small difference
between two large strings is considered more similar than the same amount of
difference between smaller strings. Formally, a normalized distance is a
metric $m(.,.)$ that takes values in $[0,1]$ and satisfies the Kraft
inequality:
\begin{equation}
  \label{eq:kraft}
  \sum_{\{ y | y \neq x\} }2^{-m(x,y)} \leq 1
\end{equation}

Whereas information distance metric $e(x,y)=max\{ H(y/x), H(x/y) \}$
\cite{information-distance} is not normalized, NID achieves normalization by
an appropriate denominator to $e(x,y)$, acknowledging that the difference of
two strings is  not absolute, but relative to their complexity (i.e. magnitude)
\cite{similarity-metric}. NID has been applied to clustering in diverse
domains such as genomics, virology, languages, literature, music,
handwritten digits, astronomy, and has reported success even in
heterogeneous domains using a variety of compressors
\cite{rudi-clustering-compression}, therefore it seems a useful tool for
machine learning research.

By \prettyref{thm:condentropy} mentioned in last subsection we can
approximate the conditional entropy since $H(t/s) \approx H(s,t) - H(s)$. Each of
the terms on the right hand side can be approximated by a compression
program from the above. Normalized Compression Distance (NCD) uses a
standard compression program such as gzip or bzip2 to approximate $d(x,y)$.
\begin{equation}
  \label{eq:ncd}
  ncd(a,b) = \frac{|C(a,b)| - min(|C(a)|,|C(b)|)}{ max(|C(a)|,|C(b)|)}
\end{equation}
where $C$ is the compression function.

In another work, Kraskov et al propose using mutual information both in
Shannon's version and Kolmogorov's version based on the same proof~
\cite{kraskov-clustering}. These studies are relevant to our problem in
that they show the versatility of Kolmogorov complexity. We shall now try to
answer if we can achieve similar feats in data mining.

\section{Abstract representations}
\label{sec:abstraction}

Before proceeding with our formulation of frequent pattern discovery from an
information theoretic perspective, it is worthwhile giving a philosophical
overview of the task. The main objective of frequency mining is to summarize
a large data set. With a suitable threshold, we obtain a smaller data set
that is representative of the most significant patterns in the data. By
means of such an abstract representation, one then achieves more specific
tasks such as discovering association rules or clustering the data.

Recent formulations of association rule mining have characterized the task
as generalization of the data. This is a necessary condition for any
successful abstraction, else what use can we imagine of an abstract
representation? According to Marvin Minsky, another way of putting this
would be the removal of unnecessary details from the
representation.~\cite{minsky6} Statistically, ``detail'' could be understood
as infrequent patterns in data, which is precisely what frequent item set
mining eliminates. Thus, a comparison of the common sense notion of
``abstraction'' and the familiar data mining task of summarization is in
order.

Let us conceive of an abstract sketch A. If this drawing is an abstraction
of a lively picture B, we expect to find the most ``important'' features of
B in A, perhaps only some of them. We would also expect to see the details,
for instance the texture, shading and colors of B to be removed in A
(assuming that it is quite abstract). In addition, we would not like to see
anything in A that does not correspond to a significant feature in B.  Some
caricatures, like those of politicians drawn in a clean generic style, may
set a good example of this kind of sensory abstraction (Note however that
some caricatures are highly stylized and will set a bad example for
abstraction). The facial features in a caricature are highly informative;
they convey much information about the facial identity of the person at a
small cost of representation. On the other hand, like any other image, the
abstract representation must be built from low-level components, which are
apparently not part of the original image. If these components, such as the
basic drawing patterns of the caricaturist, are kept simple enough, the
resulting work will look abstract.

If we are to relate the above characterization of abstraction to data
mining, the most problematic part might be the ``important'' term. After
all, an important feature for one task might be unimportant for another.
Consider the notes of a symphony. The pitch and duration information is
considered significant because it helps us to quickly discern one piece of
music from the other. This is true for any given application domain. For
recognition of music, it is the pitch or the interval that matters. But for
speech, it is the phoneme that matters. The truly generic summarization
algorithm might be able to discover the concept of note or phoneme merely by
looking at the data. If we take B to be only one datum in a data set, we
will find it more productive to think of the importance of a feature
determined by the frequency of its occurrence. This approach suggested also
in the beginning of the section does not completely solve the problem,
however. We also need universal and objective criteria for determining if a
feature approximately occurs in a given datum.

Let us now make our explanations more precise. We say that A is an abstract
representation of B if and only if:
\begin{enumerate}
\item A is substantially less complex than B.
\item Every important feature of A is similar to an important feature of B.
\end{enumerate}

Note that condition 2 can also be stated as: ``There is no important feature
in A that is not similar to an important feature in B''.

This definition is more relevant to abstraction than lossy compression.
Especially, in lossy compression the only purpose is to reproduce the data
set with a low error rate (e.g. defined in terms of how well the
reproduction is), it does not necessarily take into account simplification
of condition 1. Neither does it address the ``similar'' predicate of the
last condition. One might decide to exclude color from the abstract
representation of a house, but in traditional image compression such choices
would not be considered. Furthermore, lossy compression does not take into
account the generalization power of the representation over an ensemble of
objects. However, in frequency mining, we can give a rigid meaning to
importance, e.g. statistically significant patterns.

If we now consider a frequent pattern discovery algorithm, we may say that
the set of frequent patterns satisfy conditions 1 and 2 to be an abstract
representation of the entire data set. A useful frequent pattern set is
smaller than the transaction set and each frequent pattern (all of them
above the given support can be said to be important) occurs in B as an
important feature. In this sense, the pattern set does not only model the
current data set, but presumably also future extensions of the data source.
(We can note here that the non-traditional statistics provided by the
frequent itemset-like computation may have use for predictive modelling in
general).


\subsection{Analysis of common objections}

An objection may be raised at this point with respect to the traditional
duality of syntax vs. semantics. It may be suggested that abstraction
crucially depends on semantics which does not seem to be mentioned in our
definition. It need not be, since semantic relations, too, may be accounted
for in the ``similar'' predicate. On the other hand, it must be reminded
that cryptic references should not in general be considered as abstract in
themselves. By abstraction, we refer to manifestly useful, generalized,
compact representations. Any cryptic representation may be conceived of as
an encrypted form of such an underlying ``successful'' abstract
representation.

\subsection{Other approaches for pattern interestingness}

Equating frequency with importance may not be the only or satisfactory way
of defining interestingness of a pattern objectively. If we go back to the
caricature example, an approach which takes the locality and statistics of
the image might be able to produce abstract features which are closer to the
common sense description of interestingness. In particular, using wavelets
may capture the locality of many data types~\cite{siebescomplex}. Compare
also the approach of non-linear PCA to image analysis (for the later task of
classification, etc.)~\cite{rosipal00kernel}.





\section{Algorithmic information and the concept of pattern}
\label{sec:aitpattern}

As noted by \cite{lintheory}, a pattern may be conceived of the shortest program that
generates a string. Otherwise, the concept of a pattern is something else
entirely in every machine learning and data mining paper.  By using bit
strings and programs, we can give an objective, and universal definition of
a pattern.  Algorithmic information theory can then be used to define
pattern operators in a way that is surprisingly close to cognitive
processes. However, at this stage of our research, we do not yet concern
ourselves with the programs, our patterns are simply bit strings for now.

In particular, information distance and normalized information distance
which were briefly covered in \prettyref{sec:related} are universal measures
of similarity that are completely independent of the application domain, and
some amazingly simple implementations have achieved success in diverse
domains and learning tasks. Our use of information theory is directly
related to the concept of information distance. We also use conditional
entropy to quantify structural difference.

\section{A general model}
\label{sec:general}

We are now going to generalize the set-theoretic definition of the classical
frequent item set mining to cover a wider range of scientific measurement.
Assume that we have samples of sensor data from a ``fixed'' instrumentation
device, for instance image data from a radioastronomy telescope examining a
certain region of the space. A better example could be seismograph data
which transmits measurements irregularly and for any number of samples.

Let transaction multi-set (set with repetition) $T = \{ x | x \in \{ 0,
1\}^*\}$ be the unordered list of observations from the same domain. Let also
a pattern $x \in \{0, 1\}^*$. We will say that an \emph{abstract} pattern $x$
occurs in datum $y \in T$ iff:
\begin{enumerate}
\item $H(x) \leq  c_1.H(y)$ (entropy reduction)
\item $H(x/y) \leq  c_2.H(y)$ (noise exclusion)
\end{enumerate}
where $0<c_1<1$ and $0<c_2<1$. We denote ``$x$ occurs in $y$'' by $x \prec y$.

in compliance with our observations about the nature of abstract
representations. Note that we have paid attention to make our conditions
scale-free, in that we want to find patterns that have only a fraction of the
datum complexity, and we want to exclude patterns which carry more than a
specified ratio of noise (w.r.t. datum). If we did not care for the
definition to be scale-free, we could use additive terms for our definition:
\begin{enumerate}
\item $H(x) \leq H(y) - c_3$
\item $H(x/y) \leq c_4$
\end{enumerate}
which seems less useful since in our definition we put no restriction on the
bit strings (as in the seismograph example). Therefore, we will stick with
the scale-free variants.

Having generalized the pattern occurrence operator in the set theoretic
definition from the subset operation to the information-theoretic
conditions, the rest of the problem definition is trivial. Let the frequency
function $f(T, x) = |\{ {x \prec y} | \quad {y \in T} \} | $. Our objective is the
discovery of frequent patterns in a transaction set with a frequency of $\epsilon$
and more.  The set of all frequent patterns is $\F(T, \epsilon ) = \{ {x \in \{0,1\}^* }
| {f(T, x)\geq\epsilon} \}$, which is guaranteed to be finite due to the entropy
reduction condition. (Note that we consider the classical definition of
Kolmogorov/Chaitin/Solomonoff complexity as mentioned in
\prettyref{sec:related}). However, the size of $\F$ can be quite large, as
in the frequent item set mining problem.

The downward closure lemma which states that the subsets of a frequent
pattern are also frequent makes the \algo{Apriori} algorithm possible in the
context of frequent item set mining \cite{agrawal93mining}. There is an
analogue of the contrapositive of this lemma for our general formulation.
Note that to simplify matters we assume a self-delimiting program encoding
such as LISP. According to Chaitin, this is not much of a requirement. (The
analysis without the self-delimiting condition would introduce an additive
logarithmic term which we would address separately.)
\begin{lemma}
  \label{lem:extension}
  If $x \notin \F(T,\epsilon)$ then $xy \notin \F(T,\epsilon)$. Less formally, any extension of an
  infrequent pattern is also infrequent.
\end{lemma}
\begin{proof}

  If $x \notin \F(T,\epsilon)$ then, $f(T, x) < \epsilon$. Let $z$ be any datum in $T$ for
  which it is not the case that $x \prec z$. Then, at least one of the pattern
  occurence conditions does not hold. We can now analyze whether an extended
  $xy \prec z$.
  \begin{itemize}
  \item Suppose that the entropy reduction condition does not hold:
    $H(x) > c_1.H(z)$. Then, $H(x,y) > c_1.H(z)$ since $H(x,y) > H(x)$.
  \item Alternatively, suppose that the noise exclusion condition does not
    hold: $H(x/z) > c_2.H(z)$. Then, it doesn't hold for $x,y$ either. $H(
    (x,y) / z) = H( y/(x,z) ) + H(x/z) + O(1) $ by subadditivity of
    algorithmic information. Since $H( y/(x,z))>0$ (since it has to be at
    least $O(1)$), then we find that $H( (x,y) / z) > c_2.H(z)$. 
  \end{itemize}
  Therefore, it is not the case that $xy \prec z$. Then, $f(T,xy) \leq f(T,x) < \epsilon$
  which entails that $xy \notin \F(T,\epsilon)$.

\end{proof}

\section{The Synthetic algorithm}

By \prettyref{lem:extension}, we are inspired to write an algorithm which
starts with a number of primitive candidate patterns and searches the
pattern space in breadth first fashion like the \algo{Apriori}
algorithm. First, let us look at the calculation of pattern occurrence
conditions.

\subsection{Approximate calculations}

Algorithmic information content is uncomputable using a universal computer.
Neither of the conditions we give are recursively enumerable. Fortunately,
that should not trouble us too much, for we can use the methods mentioned in
\prettyref{sec:ncd} to approximate these uncomputable values. However, it is
arguable whether using a dictionary-based simple compressor is sufficient
for the range of data mining applications we are interested in. At the
present, the only obvious advantage of using a traditional compressor would
seem to be efficiency.

We again approximate the conditional entropy using subadditivity of
information $H(t/s) \approx H(s,t) - H(s)$.  With a compressor $C(\cdot)$ such as
gzip, our two conditions become:
\begin{enumerate}
\item $C(x) \leq  c_1.C(y)$ (entropy reduction)
\item $ C(x,y) \leq  (1 + c_2).C(y)$ (noise exclusion)
\end{enumerate}

\subsection{BFS in pattern space}

We will adapt a generate and test strategy similar to \algo{Apriori} for our
first algorithm, applying the theory introduced in the paper. The pruning
logic is quite similar, we do not extend infrequent patterns by
\prettyref{lem:extension}. We will keep the algorithm as close as possible
to \algo{Apriori} to show the relation, although there could be many
efficiency improvements following various frequent itemset mining
algorithms. \algo{Micro-Synthetic} extends the pattern length by $n$ bits at
each iteration of the algorithm. Initially, a fast algorithm finds all
frequent patterns up to $n$ bits (akin to discovery of large items). The
\algo{Generate} procedure extends the frequent patterns of the previous
level up to $n$ bits. Then, a database pass is performed and the pattern
occurence conditions are checked for each candidate pattern and transaction
element. Then, the algorithm iterates, generating candidates from the last
level of frequent patterns discovered, until we reach a level where there
are no frequent patterns, exactly as in \algo{Apriori}.


\begin{algorithm}
  \begin{small}
  \caption{ $\algo{Micro-Synthetic}(T, \epsilon, c_1, c_2)$}
  \label{alg:synthetic}
  \begin{algorithmic}[1]
    \STATE $F_0 \gets \{ |x| \leq n | x \in \{0,1\}^* \land f(T, x) >= \epsilon \} $
    \STATE $k \gets 1$
    \WHILE{ $F_{k-1}\neq\emptyset$}
    \STATE $C_k \gets \algo{Generate}(F_{k-1})$
    \FORALL{$y \in T$}
    \FORALL{$x \in C$}
    \IF{$C(x) \leq  c_1.C(y)$ $\land$ $ C(x,y) \leq  (1 + c_2).C(y)$}
    \STATE $count[x] \gets count[x] + 1$
    \ENDIF
    \ENDFOR
    \ENDFOR
    \STATE $F_k \gets \{ x \in C_k | ~ count[x] \geq \epsilon \}$
    \STATE $k \gets k+1$
    \ENDWHILE
    \RETURN $\bigcup_k F_k$
  \end{algorithmic}
  \end{small}
\end{algorithm}

\section{Discussion}

The algorithm is called \algo{Micro-Synthetic}, because direct search in
pattern space has obvious limitations. On the other hand, that is also what
all frequent pattern discovery algorithms do, therefore it may not be at a
greater disadvantage. Like in the basic frequent itemset mining algorithms,
the support threshold must be given. However, we also require two extra
parameters to delimit the pattern occurence. Unfortunately, our formulation
falls short of the ``parameter-free'' ideal \cite{paramfree}. At the moment,
we can give no guidelines for setting $c_1$ and $c_2$ except that they must
be small enough. Especially $c_2$, which controls vagueness in our model.
The basic frequent item set mining problem has no place for vagueness, the
pattern relation is strict. On the other hand, our formulation places no
bounds on the kind of data/pattern representation, and allows for vague
representations, which are useful for a system that can abstract.

An implementation effort is ongoing. \algo{Micro-Synthetic} has been implemented
and tested on small datasets. We have tried a variety of compressors
like gzip, bzip2 and PAQ8f for the information distance approximation. 
While we have managed
to find some interesting character patterns this way (such as finding an abstract
pattern of 00000001111111 from example strings of different length which contain a sequence of 0's and 1's in them, with errors), we have observed that the
suboptimality of the compressors (relative to the particular decompressor) 
causes too many random patterns to be found, which cannot be attenuated by
the $c_2$ parameter. We have been thus working on a 
simple but optimal compressor that will fit out implementation better. After
we get some results using the \algo{Micro-Synthetic} on toy problems, 
we are planning to devise an algorithm with many optimizations to deal with more
realistic data sets.
We think that an implementation could demonstrate results on
both traditional tabular datasets, and novel kinds of data due to the
generality of data schema, depending on the availability of a suitable compressor.

\section{Conclusions and Future Work}

We have made a high-level analysis of the frequent pattern discovery
problem, by observing relations between the common sense notion of
``abstraction'', and the summarization task. We have determined objective
criteria for a pattern to be an abstract representation. These criteria were
interpreted as information theoretic conditions of reduced entropy and noise
exclusion for a problem definition where patterns and data are any
bitstring. We have replaced the pattern occurence operation in frequency
mining with the conditions we have proposed. Thus, we have achieved a
generalized version of the frequent pattern discovery problem. Thereafter,
we have demonstrated that our conditions allow for pruning which is
essential for the search in pattern space, which is vast but bounded. We
have then used commonly employed methods to apply Kolmogorov complexity in
real-world to design an algorithm suitable for the discovery task. Finally,
We have introduced an \algo{Apriori} like algorithm which enumerates all 
frequent patterns in our formulation.

Our research requires yet a lot of work to be done, both in the theory and
experimental studies. First, there should be theoretical properties
that should be clarified, and alternative search methods should be
analyzed. Especially, pattern space clustering methods and efficient
representations may be sought. We have given an algorithm only for all
frequent pattern discovery, the analogues of closed/maximal mining may be
investigated. Second, a synthetic data set generator should be written, which
highlights the virtues of our model and if possible real-world data should
be tried out. Third, the effects of different kinds of compressors must be
analyzed.

\section{Acknowledgements}

Many thanks to anonymous reviewers who spotted problems with an initial 
version of this paper. Thanks to Matt Mahoney for making the latest PAQ code
available. 

\bibliography{agi}
\bibliographystyle{aaai}

\end{document}